\documentclass{article}

\usepackage{fer}

\usepackage[utf8]{inputenc} % allow utf-8 input
\usepackage[T1]{fontenc}    % use 8-bit T1 fonts
\usepackage{hyperref}       % hyperlinks
\usepackage{url}            % simple URL typesetting
\usepackage{booktabs}       % professional-quality tables
\usepackage{amsfonts}       % blackboard math symbols
\usepackage{nicefrac}       % compact symbols for 1/2, etc.
\usepackage{microtype}      % microtypography
\usepackage{lipsum}		% Can be removed after putting your text content
\usepackage{graphicx}
\usepackage[square,sort,comma,numbers]{natbib}
\usepackage{doi}
\usepackage{floatrow}
\usepackage{enumitem}
\usepackage{geometry}

\title{AI-based facial emotion recognition solutions for education: A study of teacher-user and other categories}

\author{ \href{https://orcid.org/my-orcid?orcid=0000-0001-9060-4702}{\includegraphics[scale=0.06]{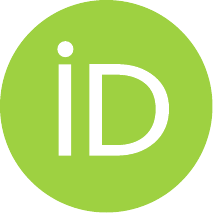}\hspace{1mm}R. Yamamoto~Ravenor}\\
	Faculty of Science\\
	Ochanomizu University\\
	2-1-1 Otsuka, Bunkyo City, Tokyo, Japan\\
	\texttt{yamamoto.ravenor@ocha.ac.jp} \\
}

\hypersetup{
pdftitle={AI-based facial emotion recognition solutions for education: A study of teacher-user and other categories},
pdfauthor={R. Yamamoto Ravenor},
}

\begin{document}
\maketitle

\begin{abstract}
	Existing information on AI-based facial emotion recognition (FER) is not easily comprehensible by those outside the field of computer science, requiring cross-disciplinary effort to determine a categorisation framework that promotes the understanding of this technology, and its impact on users. Most proponents classify FER in terms of methodology, implementation and analysis; relatively few by its application in education; and none by its users. This paper is concerned primarily with (potential) teacher-users of FER tools for education. It proposes a three-part classification of these teachers, by orientation, condition and preference, based on a classical taxonomy of affective educational objectives, and related theories. It also compiles and organises the types of FER solutions found in or inferred from the literature into \textit{technology} and \textit{applications} categories, as a prerequisite for structuring the proposed \textit{teacher-user} category. This work has implications for proponents', critics', and users' understanding of the relationship between teachers and FER.
\end{abstract}

\vspace{1\baselineskip}


\vspace{1\baselineskip}

\section{Introduction}
The overwhelming bulk of literature about artificial intelligence (AI) in schools is about AI on students. The very idea that there is a question about AI on teachers is largely unexplored. Although the fact is flagrant and undeniable, it may be worth a moment’s examination. For instance, since the beginning of the Artificial Intelligence in Education series from Springer's Lecture Notes in Artificial Intelligence, on average, the word "teacher/s" counts over nine times less than the word "student/s", as if teaching and learning have lost touch with each other.

There is also no need to prove that the teaching art, unlike the teaching technique, is not supported by mechanistic constancy and ratiocination, but rather by improvisation and sub-rational expedients, which is why education depends upon teachers, not on machines. The art of teaching, like the art of parenting, is one of the most complicated activities human beings do, and, because it involves far more emotions than we realize \citep{plutchik2013emotions}, neither humans nor machines can truly be trained for it. 

Although numerous machines can and do perform teachers’ roles in terms of teaching techniques and evaluating student cognition \citep{davies2017programmed}, existing “emotional” machines designed to replicate elements of the teaching art are more like sensory and analytic “prostheses” for teachers rather than functional replacements. This paper is concerned with one subset of these teaching artefacts, the AI-based facial emotion recognition tool.

Facial emotion recognition (FER), facial expression recognition, facial affect detection etc. are terms often used interchangeably to refer to a technology or methodology designed to detect sentiment cues from the face. In plain terms, on the one hand, FER amplifies visual details just like a magnifying glass or binocular, and on the other, it acts like a translator, converting facial descriptions from mathematical language into another language. By enhancing human natural ability to recognise and analyse facial characteristics it opens up a wide range of applications such as security, authentication, and emotion detection. Published research discussion on the application of automatic facial emotion recognition in education (henceforth “FERed”) begun perhaps with \citep{pantic2007machine}. As when some teachers, though not all, observe their students’ facial expressions to formulate a provisional hypothesis of the teaching-learning outcome, so a FER system can “watch” the students or recordings of them, collecting and processing facial expression-related data. The technology needed is now already in place, but the standards are not.

The proposition that an element of a school population should be subjected to FER experimentation without declaring exactly whose efficiency it enhances, with what scope, and by what means, seems insensitive, to say the least, as its critics have pointed out \citep{andrejevic2020facial, mcstay2020emotional, williamson2020datafication}. By the same token, to criticise FERed without specifying exactly the points of concern, seems futile. 

Studies on FER in education, as far as examined by this study, are characterised in terms of technology types and/or applications. This study introduces the category of “users” to call due attention to variations in teacher-user types, needs and wants, which it attempts to determine based on established theories in education and cognate sciences. In doing so, it touches on several questions prudently avoided in the literature about FER in education.

\section{The student face and learning}
\label{sec:subsection{Krathwohl’s taxonomy of affective educational objectives}}

\subsection{Question \#1. What is a “learning” face?}

Some of the most erudite teachers in history, such as Pythagoras and Socrates, are known to have taken student physiognomy very seriously in distinguishing those who \textit{can} learn from those who cannot \citep{liggett1974human}, but no equivalent evidence was found concerning teachers’ efforts to identify students who \textit{are} learning from those who are not. In recognizing, for instance, that someone is in pain, health professionals consider the facial expression a more reliable source of information than a patient’s verbal account of the pain \citep{craig2011facial}. While the recognition of a “suffering” face and that of the “learning” face may be related, no studies were found to either define a “learning” face, or to ascertain the extent of teachers’ reliance on it. 

Because everyone has facial habits and characteristics, one can make better sense of emotions on a learner’s face \textit{after} becoming familiar with the framework of that face and the ways in which that framework is used when learning. A contraction and furrowing of the brow may be: \par
a) a sign that the learner is concentrating, or confused, or angry etc. \par
b) simply a personal habit one needs not explain or apologize for.\par
When a teacher claims to have detected certain emotions on a student’s face, credibility largely depends on one’s subjective perception of the teacher making the assertion. But when those making the assertion are scientists, their speculations are typically produced in a form which can be mistakenly taken as authoritative. 

\subsection{Question \#2. What emotions are compatible/incompatible with learning?}

First, despite the many theoretical propositions suggesting that positive emotions increase the propensity for learning \citep{fredrickson1998cultivated, pekrun2002positive, Pritchard2003UsingEA, schutz2006emotions, ryan2018self, carmona2019psychological, camacho2021activity}, there is no criterion whereby some emotions are considered conducive to learning and others not. Let us imagine that student X, whose parents are bankrupt, takes a course on banking. While X’s emotions may all be negative, the desire to learn in this course may nevertheless be at its apex. By the same logic, student Y, being happily in love, constantly manifests positive emotions but is unable to focus on learning. However crude such examples may be, the duty to recognise and refute convenient categorisation of emotions as “learning-compatible” and “learning-incompatible” should be assumed. Those who made Pekrun's work one of the most cited sources of confidence in categorical emotions for FERed, may have missed the following words "simplistic conceptions of negative emotions as bad and positive emotions as being good should be avoided because positive emotions are sometimes detrimental and negative emotions such as anxiety and shame beneficial." \citep{pekrun2002positive}.

Secondly, there are criteria by which some emotions are deemed desirable or undesirable in schools. For instance, there is obviously something emotionally wrong with a student crying in school, just as it is inappropriate to burst into laughter during a class, if nothing is amusing. However, unless behavioural correction, psychological therapy etc. are included the education service, educators are not strictly and directly concerned with student conduct, depression, maladjustment etc. 
Emotions like anger, sadness, shame, fear etc. can greatly motivate learning; frustration, rage, disobedience etc. are not related with ignorance; what may seem boring for some students can be exciting for others; disengagement with one subject or activity may be due to intense engagement with another; and the list of complexities can go on. One only needs to briefly examine the lives of people like Isaac Newton, Marie Curie, Mahatma Gandhi or Albert Einstein to grasp the complexity and intricacy of the connections between emotions, behaviors, and achievements.

\subsection{Question \#3. How can machines distinguish general emotions from learning-related emotions on student face?}

Student affect can be related either to general emotional state, or to emotional response towards the educational content. Although good practitioners of the teaching art, as \citep{durkheim1973moral} would call them, may intuitively perceive or infer the difference between the two through subconscious processing of accumulated experiences, there are no known ways of making the distinction between general and particular emotions accessible from a mathematical standpoint so that it can be done by machines.

Many FER proponents suggest that FER-generated feedback on student affect can serve as basis for teachers to implement personalised and/or generalised interventions \citep{mazza2007coursevis, martinez2012interactive, aslan2019investigating, ngoc2019computer, ashwin2020automatic, wang2020emotion}. \citep{lin2020continuous} presupposed that “academic emotions”, as defined by \citep{pekrun2002positive}, can be recognized on a student’s face by a machine, and went on to develop a FER system that reportedly identifies such emotions of students using a model for continuous facial emotional patterns. 

The assumption that academic emotions can automatically be detected has gone as far as to call for facial expression databases focusing on academic emotions. One such example is DAiSEE, reported to have been used in seven studies \citep{lek2023academic}. Researchers who work with this assumption seem to equate \citep{pekrun2011measuring}’s psychological methods of measuring academic emotions, and differentiating them from general emotions, with mathematical methods of observation and analysis used in FER. 

Let us, nevertheless, assume that FEReds can distinguish between general and academic emotions. If FEReds designed to detect only general  emotions on each student’s face were used by teachers in schools, it follows that teacher-users are both willing and able to adequately respond to FER-generated negative feedback on student affect. In reality, however, the typical teacher is neither trained nor constituted as a mental health professional or as an entertainer. Teachers are concerned only with certain aspects of the personal and intellectual development of their students and, in that role, each student’s general emotional state is a datum from which they start. It is neither a teacher’s duty to change it, nor reasonable to attempt during teaching hours for it is changeable within measurable time. Teacher intervention to change student general emotions, not only goes beyond the scope of standard education, but may also do more harm than good. Therefore, if FEReds are designed to detect general emotions, then they may be more suitable for categories of users other than teachers.

Conversely, if FEReds programmed to detect student emotions related to the educational content before them, it follows that teacher-users can make sense of the feedback received from the machine. Many teachers assume the moral duty to creatively intervene in order to elicit and maintain students’ interest, cultivate their innate emotional intelligence etc., in freedom, by trial and error, not in textbook fashion or by imposition. Such teachers may find useful a FERed that offers insight into student emotional responses, provided these are aligned with affective educational objectives that teacher-users themselves are familiar with. The details of these objectives vary with theoretical persuasions, teaching experiences, educational traditions in a teacher’s country of origin, the teacher’s age etc. All these variations cannot be explored here, but section 3.1. \ref{sec:subsection{Krathwohl’s taxonomy of affective educational objectives}} touches on the first of them, giving an account of attributes contained in a classical theory of affective learning behaviour is included in section. 

The considerations above invite efforts attempting to automatically detect student emotions using FER to: \par
a) be guided less than some have been \citep{ashwin2020impact, bouhlal2020emotions, lin2020continuous, Li2021, ramakrishnan2021toward, sumer2021multimodal, savchenko2022neural, luo20223dlim, chen2023stran, naga2023facial}, by oversimplifications which a machine translates into categories (such as compatible with learning and incompatible with learning) and more by empirical insights directly from education practitioners. \par
b) acknowledge that the distinction between general and education-related emotions can hardly be hoped to formulate as an algorithm; and \par
c) declare in which of the two broad and distinct streams of professional practice, i.e. educational and non-educational, their work lies.

\section{Traditional evaluation of affective educational objectives}
\label{sec:section{Traditional evaluation of affective educational objectives}}

While student cognitive behaviour is considered public matter, the affective one is deemed private. This is one reason why, despite undeniable affective implications of all teaching and learning, typical schools are only preoccupied with the evaluation of cognitive educational objectives (cognitive testing), relegating the evaluation of affective educational objectives (affective testing) to the status of a teacher's hobby, for which there is no responsibility to expend public resources on. Another reason is that, unlike cognition, which is adapted to quantitative analysis, measuring student affective achievement is difficult, whether the method is scientific or not. 

\subsection{Krathwohl’s taxonomy of affective educational objectives}

Bloom \citep{bloom1956taxonomy} brought out the distinction between cognitive and affective domains of educational objectives, pointing to a scale of consciousness on which the latter is positioned lower than the former. Krathwohl's \citep{krathwohl1964taxonomy} affective taxonomy deals with objectives expressed as interest, attitudes, values, appreciation and adjustment, which are tested for using questionnaire strategies. Given the wide meaning of these terms, they were encompassed into ranges of behaviour and ordered into five categories. \par
1.0 Receiving \par
2.0 Responding \par
3.0 Valuing \par
4.0 Organization\ par
5.0 Characterization by a value complex \par
\begin{figure}[!htb]
    \centering
    \ffigbox[\FBwidth]
    {
        \includegraphics[height=14.81cm, width=10.41cm]{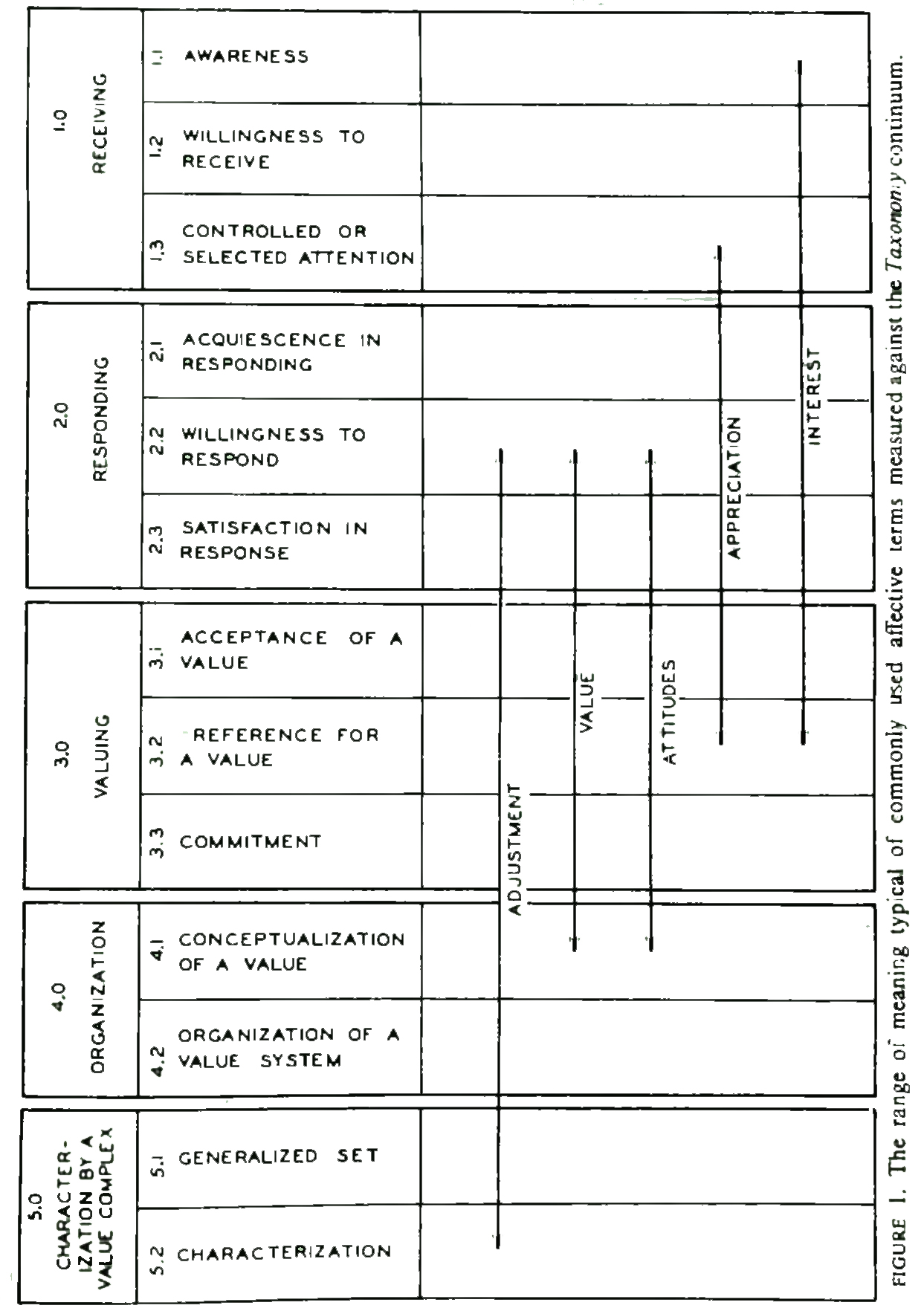}
    }
    {
        \caption{The range of meaning typical of commonly used affective terms measured against the Taxonomy continuum. \citep[p.37]{krathwohl1964taxonomy}}
        \label{fig:range}
    }
\end{figure}

The lowest level in the affective continuum is characterized by a state in which the student is attentive and passively “receives” the teaching. \textit{Acceptance} is an active state marked in the “responding” level. At the third level, the student already pursues the subject or activity. At the fourth and fifth levels, the behaviour is described as attitudes form a structure within a network of values. In the range of meaning, \textit{interest} can be located between the starting phase of “1.0 Receiving” and the middle of “3.0 Valuing”. \textit{Appreciation} overlaps with \textit{interest} to a greater extent than \textit{attitudes}, \textit{values} and \textit{adjustment}, which are marked at higher levels in the taxonomy. Krathwohl et al. do not categorise emotions as more or less conducive to learning, but do remark that behaviour with emotions such as enthusiasm, warmth, disgust, etc. is little at the lowest levels of the taxonomy, but prominent at the middle levels, and decreased towards the top \citep{krathwohl1964taxonomy}. Figure \ref{fig:range}, taken from \citep{krathwohl1964taxonomy} depicts where student interest and the other goals are located. \footnote{A replication of this figure was omitted to avoid truncation.}

The original affective taxonomy elaborated by Krathwohl et al. inspired and provoked numerous researchers to produce other such taxonomies \citep{hoepfner1972cse, brandhorst1978reconceptualizing,gephart1980evaluation}  and related classifications \citep{mccombs1990putting, clarkson2010mathematics}. Of all the affective taxonomies noted, Krathwohl’s remains the most prescriptive and its limitations are thoroughly recognized by the authors. One such limitation is that the objectives expressed in terms of values, attitudes etc. are not well operationalized and thus the taxonomy is recommended for curriculum construction rather than instruction planning. Another limitation stems from the difficulties that authors admit facing when making the distinctions between and among categories. Some critics of Krathwohl’s taxonomy pointed out that the concepts used, such as the division of affective activities into "receiving" and "responding" categories, is strongly based on behaviourism\citep{yun2021affective}. Others remarked that Krathwohl’s affective taxonomy focuses on internal constructs, which goes against the behaviourist focus on observable behaviours \citep{irvine2021taxonomies}. Although admittedly too general, abstract and limited in scope, Krathwohl’s taxonomy remains the landmark in the affective domain of educational objectives.

\subsection{\textit{Student engagement} as an affective learning objective}

Krathwohl’s taxonomy did not refer to “engagement”, but over time the word “engagement” gained popularity in works related to the affective domain of educational objectives. \citep{fredricks2004school}, drawing on the taxonomies of Bloom \citep{bloom1956taxonomy} and Krathwohl \citep{krathwohl1964taxonomy}, identified three dimensions to student engagement: \par
a) “behavioural engagement”, which implies that the student is present, attentive, participatory \rule[0.5ex]{1em}{0.55pt} similar with the first phase of Krathwohl’s “1.0 Receiving” category, where interest may be covert, extrinsic, or passive (such as when a student is: either not necessarily interested in what is being taught, but wishes to make a good impression, obtain praise, high grades, degrees etc.; or ready to become interested in the subject or activity, though not interested yet); \par
b) “emotional engagement”, characterised by affective reactions like interest, enjoyment, sense of belonging \rule[0.5ex]{1em}{0.55pt} corresponding with Krathwohl’s “2.0 Responding” and “3.0 Valuing” categories, where interest is overt, active and intrinsic (as when a student is more interested in the educational content than in getting good grades); and \par
c) “cognitive engagement”, describing students invested in learning and willing to go beyond requirements \rule[0.5ex]{1em}{0.55pt} Krathwohl recognizes that the last two categories “4.0 Organization” and “5.0 Characterization by a value complex” in his affective taxonomy are, at least in part, cognitive (student conceptualizes the value to which he previously responded, and this value is integrated and organized into a value-system which may come to characterize the student as an individual).

\subsection{\textit{Student interest} and \textit{student engagement}}
John Dewey remarked that \textit{interest} “has its emotional as well as its active and objective sides” and that “the root idea of the term seems to be that of being engaged, engrossed, or entirely taken up with some activity because of its recognized worth“ \citep{dewey2022collected}. In other words, to say that a student is “interested” means that the student is either engaged, absorbed, or consumed by whatever is that interesting to that student. \citep{renninger2015interest} wrote: “Interest is one indication of emotional engagement”. While Dewey acknowledges that interest may not necessarily imply engagement, Renninger seems to suggest that whenever someone is interested in something, emotional engagement is always present. To further complicate matters, one may consider \citep{freeman1998interest}’s assertion that “engagement is only possible when interest is present”.

This study admits that a student can be genuinely interested in something without being actively engaged in it, and does not take into account “emotional engagement”, which it regards as passive behaviour, likely undetectable and, thus, irrelevant for the intended purpose. It also considers that engagement on the part of a student does not necessarily indicate interest, as it may be when the student’s motivation may be extrinsic, rather than intrinsic (i.e. student behaviour may be influenced more or entirely by desires other than learning), a distinction that a machine cannot make. Finally, this study takes \citep{krathwohl1964taxonomy}’s view that \textit{interest} is a primary educational objective, and argues that while genuine interest (marked in Krathwohl’s affective continuum by transition from passive to active responses) might be detected using a combination of FER techniques and teacher vigilance, genuine engagement (located past subcategory “3.3 Commitment”) may not.

\subsection{\textit{Verbal inquiry} and \textit{visual detection} as student affect assessment methods}

Traditional theories on student affect, including the ones mentioned above, typically rely on \textit{asking} students to answer questions related to their feelings. Here, this method is called “verbal inquiry” (or “inquiry”) to contrast it with “visual detection” (or “detection”), as commonly practiced in fields like psychiatry and criminology. 

Logically, there is no need to inquire if something is known, just as it is futile to inquire about something which is not knowable. The method of asking is practical when the subjects are \textit{not} well known. If the subjects are well known, observation is natural. A distinction can thus be made between: \par
a) “Detective” teachers \rule[0.5ex]{1em}{0.55pt} Teachers who know their students very well would not need to ask them questions about their feelings as much as those who just met their students, and are likely to try and detect their students’ emotions \par
b) “Inquisitive” teachers \rule[0.5ex]{1em}{0.55pt} Teachers who are not (yet) well acquainted with their students may rely less on visual detection and more on asking students for information, which students are, in fact, under no obligation to supply if not willing.

\subsection{Detection of student \textit{attention}, \textit{interest}, \textit{satisfaction} and \textit{emotional intelligence}}

Four aspects extracted from \citep{krathwohl1964taxonomy}’s work are especially relevant in the context of student affect detection: \par
a) \textit{Student attention} is \textit{not} an indication of genuine interest in the educational content. \par
The “1.0 Receiving” category is characterised by an “extremely passive position or role on the part of the learner” whereas subcategory “2.1 Acquiescence in responding” is considered “the first level of active responding after the learner has given his attention”. Because many approaches \citep{ashwin2020impact, goldberg2021attentive, sumer2021multimodal, chen2023stran} to testing student affect, both traditional or AI-based, focus on seeing that each student is attentive or “receiving” the teaching, to which the present study adds that \textit{receipt} does not constitute \textit{acceptance}. The student may examine the received content at a later time and decide then whether to accept or reject it. Thus, \textit{acceptance} goes beyond \textit{receipt}, and it is the prerequisite for any advancement from the "1.0 Receiving" level to the level in Krathwohl’s taxonomy, which a student reaches after transitioning from a passive to an active stance. Acceptance or refusal is conditioned by manner and time, variables whereby student's affective state can be tracked and measured. \par
b) \textit{Student interest} is considered overt emotion, thus detectable only before \textit{values} become internalised. \par
The range of overt emotions coincides with that of interest, both being commonly observed at the middle levels of the affective continuum, which denotes the probability for detection of a student's interest, or readiness to become interested. While a student's interest may occur in the “1.0 Receiving” level (i.e. the student giving passive attention to the subject), emotion is considered covert and undetectable at that level. Emotion decreases as interest becomes internalized, and may not be detected beyond the “3.0 Valuing” level. If the idea that interest stops occurring when a student becomes committed to a particular subject seems baffling, it is because, in the affective taxonomy, “interest” is understood as a process rather than as an outcome. \par
c) \textit{Student satisfaction} may or may not be detectable. \par
In Krathwohl’s words “Emotional responses, even those that signify satisfaction and
enjoyment, may not necessarily be overtly displayed.” \par
d) \textit{Emotional intelligence} is considered covert emotion, thus undetectable \par
Emotional intelligence (according to \citep{goleman2020emotional}, emotional intelligence includes one's ability to control and motivate oneself, zeal and persistence) seems to be located along categories “4.0 Organization” and “5.0 Characterization by a value complex”, which evaluate affective objectives that “appear to require, at the very least, the ability to (…) comprehend” \citep{krathwohl1964taxonomy}.

\section{FER as an automated visual detection method}

The student’s face has been a repository of emotions pertaining to educational achievement since the beginning of teaching, but until the advent of FER it has hardly been considered analytically decipherable. Recent studies in affective computing and AI propose a variety of solutions for this art of detection, or of guessing, whichever may be the case. A variety of works on education-related FER solutions can be found in both scholarly and scientific literature, varying, as far as is known, only in terms of technology types and application-related criteria. This section reviews these two broad categories, recognising that the listing is by no means definitive or exhaustive, its purpose being solely that of expanding discussion about FERed teacher-users, the meaningfulness of which is heavily dependent on understanding existing classifications. 

\subsection{Categorisation patterns by technology type}

There are many different categorisation studies related to FER, in general, \citep{martinez2017automatic, baskar2018facial, bhattacharya2019survey, li2020deep, patel2020facial, ullah2020systematic, revina2021survey, canal2022survey, maithri2022automated, naga2023facial} and \citep{lek2023academic}'s review of FER for education, in particular. Because the technology behind FER is common to all sectors of application, the listing provided here makes no distinction between FER and FERed in terms of the technology available. This categorisation is designed to help non-specialists navigate through existing and emergent types of FER, grouped into: methodology, implementation, analysis and ownership.

\section*{}

{\footnotesize % Set the font size to footnote size

A. Methodology
\begin{itemize}
  \itemsep0em
  \item[] A.1. Algorithms
  \begin{itemize}
    \itemsep0em
    \item[] A.1.1. Traditional machine learning pipeline
    \begin{itemize}
      \itemsep0em
      \item[] A.1.1.1. Pre-processing (face detection and localisation, dimension reduction and normalisation)
      \item[] A.1.1.2. Feature extraction (e.g. local binary patterns (LBP), histogram of oriented gradients (HOG), facial landmark detection)
      \item[] A.1.1.3. Traditional machine learning (e.g. Support Vector Machines (SVM), Random Forest (RF), k-Nearest Neighbors (k-NN), Gaussian Mixture Models (GMM))
    \end{itemize}
    \item[] A.1.2. Deep learning (e.g. Convolutional Neural Networks (CNN), Recurrent Neural Networks (RNN), Long Short-Term Memory (LSTM))
    \item[] A.1.3. Decision fusion (e.g. combining results from multiple classifiers, weighted voting and probabilistic fusion)
    \item[] A.1.4. Uncertainty estimation (e.g. confidence scores and uncertainty quantification, identifying ambiguous expressions)
  \end{itemize}
  \item[] A.2. Data
  \begin{itemize}
    \itemsep0em
    \item[] A.2.1. Data collection (e.g. in-laboratory, controlled settings, in the wild)
    \item[] A.2.2. Data modality
    \begin{itemize}
      \itemsep0em
      \item[] A.2.2.1. Image/video capture hardware (e.g. webcams, smartphone, cameras, RGB/infrared/depth cameras, 3D scanners, eye-tracking devices, wearable devices) 
      \item[] A.2.2.2. Input data (e.g. image-based, video-based)
      \item[] A.2.2.3. Dimensionality (2D, 3D)
    \end{itemize}
    \item[] A.2.3. Facial expression databases (e.g. CK+, JAFFE, AffectNet, RaFD, MMI, FER2013, TFD, Multi-PIE, SFEW, Oulu-CASIA, MUG, EMOTIC)
    \item[] A.2.4. Synthetic data  
  \end{itemize}
  \item[] A.3. Data anonymisation/encryption
  \begin{itemize}
    \itemsep0em
    \item[] A.3.1. End-to-end
    \item[] A.3.2. Homomorphic
    \item[] A.3.3. Secure multiparty computation
    \item[] A.3.4. Differential privacy
  \end{itemize}
\end{itemize}

B. Implementation
\begin{itemize}
  \itemsep0em
  \item[] B.1. Connectivity (e.g. online, offline)
  \item[] B.2. Integration (e.g. unimodal, multimodal)
  \item[] B.3. Flexibility of model (e.g. adaptive, fixed)
  \item[] B.4. Real-time processing (e.g. low-latency processing for real-time applications, optimization for resource-constrained devices)
  \item[] B.5. Timing of analysis (e.g. real-time, post-analysis)
\end{itemize}

C. Analysis   
\begin{itemize}
  \itemsep0em
  \item[] C.1. Emotion interpretation 
  \begin{itemize}
    \itemsep0em
    \item[] C.1.1. Categorical approach (e.g. isolated emotions, positive, neutral, negative)
    \item[] C.1.2. Multiple dimensions (e.g. congregated emotions, arousal-valence)
  \end{itemize}
  \item[] C.2. Visualization (e.g. facial landmarks, heatmaps, AU activation maps, facial expression morphing, LIPnet, 3D facial models)
  \item[] C.3. Duration of analysis (e.g. short-term, long-term)
  \item[] C.4. Temporal context (e.g. static, dynamic, continuous/conditional monitoring)
  \item[] C.5. Scope of analysis (e.g. localized, global, hybrid)
\end{itemize}

D. Ownership and access (e.g. proprietary, open source)

} % End of footnote-size font.

\vspace{2\baselineskip}

Obviously, the choice between FERed technological approaches ultimately depends on resources and requirements.

\subsection{Categorisation patterns by technology application in education}

Although the present study did not identify any FER taxonomies based on technology application in the field of education, most proposals for FERed, mentioned earlier in this paper, are accompanied by statements regarding the application for which they were designed. It is beyond this paper’s scope to provide an exhaustive list of all possible types of FER education-related applications, the purpose being simply to contour the application grounds for FER in the realm of education.  The contrasts listed below are patterns found in or deduced from the literature, as well as models that are not covered by other studies, listed in no particular order. 

\vspace{1\baselineskip}
{\footnotesize
\begin{description}
    \itemsep0pt % Remove vertical spacing between items
    \item[A. Student target] (e.g. individual, collective, selective)
    \item[B. Educational level] (e.g. pre-school, primary education, secondary education, tertiary education)
    \item[C. Class format] (e.g. synchronous, asynchronous, online, offline, hybrid)
    \item[D. Teaching/learning approach] (e.g. traditional, interactive, adaptive)
    \item[E. Emotion focus] (e.g. general emotions, learning-related emotions)
    \item[F. Affective objective] (e.g. student discipline / attention / well-being / satisfaction / engagement / interest / emotional intelligence)
    \item[G. Content creation] (e.g. emotion-driven content creation, curriculum customisation, educational design / guidance)
    \item[H. Assessment improvement] (e.g. assessment enhancement / accuracy / personalisation, traditional / adaptive assessment)
    \item[I. Various accommodation] (e.g. support for special needs, inclusive education)
    \item[J. Adoption model] (e.g. top-down, bottom-up)
    \item[K. User target] (e.g. user-specific, generalised)
\end{description}
}

\vspace{1\baselineskip}
This kind of categories collectively shape the decisions for applying FER in educational settings, considering the unique requirements and goals of each scenario.

\subsection{Question \#4. How does the model for FER adoption (top-down/bottom-up) in educational settings impact the way FERed is made?}

Existing FERed-related research, which naturally started from the assumption that there is demand for FER in education, make no allusion to either types of users or user needs. Although users are generally presumed to be teachers, the envisioned model for FER adoption in educational settings is top-down, as if nothing could or should empower teachers to have, by their own accord, a FERed in the classroom. If research on FERed would focus on teachers’ demand for detection first, and emotion recognition second, it may avoid being ridiculed as “a solution in search of a problem” \citep{selwyn2020postdigital}, or understood as a imposition from on high that teachers can neither have a say in or refuse. It is a fact that, at least in Japan, education administrators decide on the adoption or rejection of AI in schools often without thorough consideration for, or consultation with teachers \citep{yamamoto2021intelligent}. The problem with the top-down approach is, first, that it takes the focus away from differences between users in general, and teacher-users in particular; and secondly, that it promotes a standardisation process to reduce the number of different FEReds used, in effect, reducing the number of different arts of teaching, instead of acknowledging and supporting teaching practice diversity.

This study rejects the top-down FERed adoption preconception as erroneous, and addresses the gaps it created in existing FERed categorisation patterns by proposing a new category (that of “users” with focus on “teacher-users”), the obscuring of which it plausibly caused. 

\section{Methodology}
Describing the relationship between FERed and (potential) teacher-users is complex because it requires a kind of interdisciplinary understanding that spans the boundaries between an exact science and an art. The fact that there are few methodological models for this kind of research only adds to the challenge. For the purpose of concluding this investigation and offering further direction for those engaging in a similar undertaking, the approach chosen is categorisation, recognising that the task is not so complex as to defy it.

The discerning of types is one of the most fundamental branches of knowledge. The method employed here involves engaging with existing literature, and aligns with the approach of incorporating underlying principles and unifying themes while organising and categorising in meaningful ways. It is designed to satisfy the need for systematic and rigorous deduction of pure notions, as well as structured and cohesive categorisation method in scholarship. Extensive literature review aided the identification of relevant criteria, factors and gaps within prior categorisations involving FERed technology and/or FERed applications, to which this study adds FERed users, to advance a classification of \textit{FERed teacher-users}. 

The previous two categorisations in this study took place when a FER was recognized as belonging to a class based on certain technology-related or application-related criteria. For instance, when a FER is described on the basis of its methodology, a category “methodology” is established as the class encompassing FER solutions that share a description by their technological characteristics. Further categorisation allowed for grouping tools that utilize comparable technological frameworks or mechanisms. Similarly, for descriptions of FERed from the viewpoint of FERed applications, another category “applications” was formed, representing FER solutions that serve particular educational purposes, organised in subcategories, such as student engagement assessment, student interest detection, student attention surveillance \citep{chen2023stran} etc. Because these were merely a compilation of classes, the majority of which can be found in literature, they seemed more fitting in the sections on literature.

As noted above, the categorisation approach in this study goes beyond these two stated or implied criteria, to introduce a new category dedicated, in general, to the users of FERed and, in particular, to teacher-users. When tools are sorted into this user-centric category of “users”, by acknowledging the different users subcategories, such as teachers, education administrators, parents, students themselves, researchers etc., allows one to distinguish tools that cater to different user needs and wants. Finally, the study hones its focus on a single but significant subset within the user subcategories, the teachers, further classifying them based on theoretical traditions in education and related fields.

\section{New categorisation}

Although understanding user needs is critical to the success of all automation, both proponents and critics of FERed seem to have omitted two fundamental questions: \par
a) What categories of FER-users exist in educational settings; and \par
b) What are the needs of the FERed end user. \par
This section presents a categorisation attempt intended to provide general help to those concerned with FERed, by highlighting the distinction between teacher-users and non-teacher users, as well as making better sense of the variety of teacher approaches to visual detection of student affect on the face.

\subsection{General categorisation by user-types }

Here, the purpose of presenting a list of potential FERed users is not only to provide context for narrowing the focus down to teacher-users, but also incipient criteria for user-centred FERed design thinking processes and standardisation initiatives. It is a broad overview of the main potential categories of FERed users which may be conceived in ignorance of their specific particularities, needs and preferences. Neither is this listing in a particular order, nor does it purport to include all conceivable categories of FERed users. It is also beyond the aim of this paper to enter into specific subcategories or provide detailed explanations of user needs for each category in this simple list.

{\footnotesize
\begin{description}
    \itemsep0pt % Remove vertical spacing between items
    \item[A. Teachers] 
    \item[B. Parents]
    \item[C. Students]
    \item[D. Researchers]
    \item[E. Evaluators]
    \item[F. Psychologists]
    \item[G. Education board representatives]
    \item[H. Policy makers]
    \item[I. School administrators]
    \item[J. Special education professionals]
    \item[K. Teacher trainers]
    \item[L. Educational technologists]
    \item[M. Counsellors]
    \item[N. Curriculum designers]
    \item[O. Talent scouts]
\end{description}
}

\subsection{Further categorisation of “Teachers” }

Skipping the most elementary step in innovation creation, that of user identification and consideration \citep{coleman2016design}, to report good results, advance the state-of-the-art etc., may be common practice in computer science research, but it may not always be good practice. At least in the case of educational technology (edtech) for teachers, there are two schools of thought, which vary in shades, but might be classified in this way (as \citep{giacomin2014human}'s human- vs user-centred framework does not seem to fit): \par
(1) \textit{For} creating new (non-human) teachers (e.g. intelligent tutoring systems, adaptive learning platforms, virtual / augmented reality)\par
(2) \textit{For} making (human) teachers better (e.g. FER, speech recognition tools, eye tracking systems, data analytics platforms, AI-based assessments)\par
The word “for” does not necessarily mean “against” the other group.

FERed is not in the category of educational tools created by ideas belonging to school (1), which may leave nothing for human teachers to do other than act as on-site support \citep{roll2016evolution, aslan2019investigating}, raising the question of whether they are still “teachers”. Tools belonging to (1) often do not require a user-centred design, so development can speed past research on users. The ideal of FERed is not only to enhance the teacher’s visual and analytical capabilities, but to make the teacher better. FERed belongs in school (2), and as such it requires a more direct focus on user types and needs. 

FERed “for teachers”, as some designate it \citep{zhang2021cross, luo20223dlim, ramakrishnan2021toward}), is not necessarily a universal application "for teachers", but can and should be related to a specific description of a teacher, or group of teachers. It is a matter of much more than semantic importance, because not having appropriate words whereby to distinguish teachers who are likely to become FERed users from those who are not,  has implications on the whole understanding and discussion of the subject. Under the difficulty of expression, some take refuge in teacher quality, and refer to “good” teachers, as if there was a universally agreed-upon definition of what a “good” teacher might be. In older publications, education scholars \citep{krumboltz1957effect, nelson1964affective} used some terms of art: \par
a) “cognitive instructors” to refer to teachers concerned with the cognitive goals of teaching; \par
b) “affective instructors”, to refer to teachers concerned with student affective adjustments too.

No new efforts to word the two types of teachers have been found in more recent literature. This may be, at least in part, due to the “erosion in the meaning and substance of affective objectives”, which Krathwohl hoped to redress, sped up as the years progressed. A Google Scholar query for “affective domain" versus "cognitive domain" in education-related journals, from the year 1960 to the present time, divided into 3 periods of 20 years each, shows a steady decline from a higher interest in the affective domain during the first 20-years period, to doubled emphasis on the cognitive one in the present time.

The series of categorisations below are not intended to establish terminology, but only to describe a variety of teacher types. Like the other classifications presented in this paper, this list focuses on contrasts and is not exhaustive. Unlike previous listings, this one includes some details.

\subsubsection{Categorisation of teachers by their relation with FERed}

It is common knowledge that teachers teach differently, which is what the principle of teacher autonomy is built upon. One teacher’s art of teaching is not (easily) reproducible, and one art of teaching is not generally applicable. Before a FERed is piloted in a school, one ought to know how many teachers there would be willingly, successfully and happily using it. These personal and professional characteristics, to which hardly anyone draws attention in studies related to computers in education, might be usefully classified based on teacher orientation, condition and preference (though some overlap), as follows:
\section*{}

A. Orientation
\begin{itemize}
  \itemsep0em
  \item[] A.1. Teaching philosophy (see Note 1)
  \begin{itemize}
    \itemsep0em
    \item[] A.1.1. Teachers interested in student affect
    \item[] A.1.2. Teachers not interested
  \end{itemize}
  \item[] A.2. FERed-related principles (see Note 2)
  \begin{itemize}
    \itemsep0em
    \item[] A.2.1. Teachers who do not oppose FERed
    \item[] A.2.2. Teachers who do
  \end{itemize}
  \item[] A.3. Opinion on automated FER methods related to education
  \begin{itemize}
    \itemsep0em
    \item[] A.3.1. Teachers who believe recognition of student facial emotions can be automated
    \begin{itemize}
      \itemsep0em
      \item[] A.3.1.1. Teachers who believe FERed can work for all age/education levels of students
      \item[] A.3.1.2. Teachers who believe FERed can work only for some
      \begin{itemize}
        \itemsep0em
        \item[] A.3.1.2.1 Teachers who teach students in the age/education level range that they believe FERed can work for
        \item[] A.3.1.2.2 Teachers who do not
      \end{itemize}
    \end{itemize}
    \item[] A.3.2. Teachers who do not believe recognition of student facial emotions can be automated
  \end{itemize}
  \item[] A.4. Ethical perspective
  \begin{itemize}
    \itemsep0em
    \item[] A.4.1. Privacy-oriented teachers
    \begin{itemize}
      \itemsep0em
      \item[] A.4.1.1. Teachers who believe data encryption methods can protect privacy
      \item[] A.4.1.2. Teachers who do not
    \end{itemize}
    \item[] A.4.2. Transparency-oriented teachers
  \end{itemize}
  \item[] A.5. Attitudes on student affective response
  \begin{itemize}
    \itemsep0em
    \item[] A.5.1. Attitudes on education-related affective response from students
    \begin{itemize}
      \itemsep0em
      \item[] A.5.1.1. Student discipline-oriented teachers
      \item[] A.5.1.2. Student satisfaction-oriented teachers
      \item[] A.5.1.3. Student attention-oriented teachers
      \item[] A.5.1.4. Student interest-oriented teachers
      \begin{itemize}
        \itemsep0em
        \item[] A.5.1.4.1. Teachers who work to elicit student interest in everything that is being taught
        \item[] A.5.1.4.2. Teachers who work to elicit student passion for one or several subjects 
      \end{itemize}
      \item[] A.5.1.5. Student emotional intelligence-oriented teachers
      \item[] A.5.1.6. Student engagement-oriented teachers 
      \begin{itemize}
        \itemsep0em
        \item[] A.5.1.6.1. Teachers interested in student (general) engagement
        \item[] A.5.1.6.2. Teachers interested in student (intrinsic) engagement
      \end{itemize}
    \end{itemize}
    \item[] A.5.2. Attitudes on student general emotional well-being
    \begin{itemize}
      \itemsep0em
      \item[] A.5.2.1. Teachers who believe that teacher intervention to improve student general emotional well-being is good
      \item[] A.5.2.2. Teachers who believe their intervention to improve student general emotional well-being may do more harm than good
    \end{itemize}
  \end{itemize}
  \item[] A.6. Teaching focus
  \begin{itemize}
    \itemsep0em
    \item[] A.6.1. Individual-oriented teacher
    \item[] A.6.2. Collective-oriented teacher
  \end{itemize}
  \item[] A.7. Views on emotion classification and detection
  \begin{itemize}
    \itemsep0em
    \item[] A.7.1. Teachers who believe that categorical emotions (negative/positive, happy/sad etc.) are indicative of student affective state related to learning
    \item[] A.7.2. Teachers who believe that emotional transitions (analysing congregated emotions) are more important
  \end{itemize}
\end{itemize}

B. Condition
\begin{itemize}
  \itemsep0em
  \item[] B.1. Familiarity with technology
  \begin{itemize}
    \itemsep0em
    \item[] B.1.2 Teachers who can use FERed
    \item[] B.1.2. Teacher who cannot
    \begin{itemize}
      \itemsep0em
      \item[] B.1.2.1. Teachers willing and able to learn
      \item[] B.1.2.2. Teachers unwilling, but able to learn
      \item[] B.1.2.3. Teachers unable to learn
    \end{itemize}
  \end{itemize}
  \item[] B.2. Adaptability
  \begin{itemize}
    \itemsep0em
    \item[] B.2.1. Teachers willing and able to adapt to any FERed
    \item[] B.2.2. Teachers who need a FERed to replicate, at least to some extent, their natural methods
    \begin{itemize}
      \itemsep0em
      \item[] B.2.2.1. Teachers willing and able to take part in FERed personalisation
      \item[] B.2.2.2. Teachers unable take part in FERed personalisation
    \end{itemize}
  \end{itemize}
  \item[] B.3. Adoption decision models 
  \begin{itemize}
    \itemsep0em
    \item[] B.3.1. Teachers comfortable with top-down approaches to FERed adoption
    \begin{itemize}
      \itemsep0em
      \item[] B.3.1.1. Teachers responsible for FERed use
      \item[] B.3.1.2. Teachers not responsible
    \end{itemize}
    \item[] B.3.2. Teachers uncomfortable with top-down approaches to FERed adoption
  \end{itemize}
  \item[] B.4. Technical literacy
  \begin{itemize}
    \itemsep0em
    \item[] B.4.1. Teachers who can program FERed (partially or entirely)
    \begin{itemize}
      \itemsep0em
      \item[] B.4.1.1. Teachers with FERed ownership preferences (open-source or proprietary FERed)
      \item[] B.4.1.2. Teachers with technical preferences 
      \item[] B.4.1.3. Teachers with hardware preferences
    \end{itemize}
    \item[] B.4.2. Teachers who can use FERed (by themselves)
    \item[] B.4.3. Teachers who need support (occasional or permanent)
  \end{itemize}
  \item[] B.5. Familiarity with students
  \begin{itemize}
    \itemsep0em
    \item[] B.5.1. Teachers who know their students very well
    \item[] B.5.2. Teachers who do not (yet) know their students very well
  \end{itemize}
  \item[] B.6. Teaching methods
  \begin{itemize}
    \itemsep0em
    \item[] B.6.1. Direct interaction (e.g. experiments, discussions, case studies, workshops, simulations, role playing)
    \item[] B.6.2. Content delivery (e.g. lectures, presentations, reading, demonstrations)
  \end{itemize}
  \item[] B.7. Class format
  \begin{itemize}
    \itemsep0em
    \item[] B.7.1. Online (synchronous, asynchronous)
    \item[] B.7.2. Traditional (conventional, flipped)
    \item[] B.7.3. Hybrid (online, offline)
  \end{itemize}
\end{itemize}

C. Preference
\begin{itemize}
  \itemsep0em
  \item[] C.1. Choice of emotion recognition methods (see Note 3)
  \begin{itemize}
    \itemsep0em
    \item[] C.1.1. Teachers who choose to use FERed
    \item[] C.1.2. Teachers who choose to rely on their own traditional (natural) emotion recognition methods
  \end{itemize}
  \item[] C.2. Disposition for adaptive teaching
  \begin{itemize}
    \itemsep0em
    \item[] C.2.1. Teachers willing and able to tailor educational experience for each student
    \item[] C.2.2. Teaching unwilling
  \end{itemize}
  \item[] C.3. Curriculum-related use
  \begin{itemize}
    \itemsep0em
    \item[] C.3.1. Teachers who intend to use FERed feedback for curriculum adjustments
    \item[] C.3.2. Teachers who do not 
  \end{itemize}
  \item[] C.4. Student assessment-related intent
  \begin{itemize}
    \itemsep0em
    \item[] C.4.1. Teachers who intend to include FERed feedback in student assessment
    \item[] C.4.2. Teachers who do not
  \end{itemize}
  \item[] C.5. Feedback preferences
  \begin{itemize}
    \itemsep0em
    \item[] C.5.1. Teachers who prefer to receive feedback regularly
    \item[] C.5.2. Teachers who prefer to receive feedback when significant change was detected.
  \end{itemize}
  \item[] C.6. Non-affective objectives
  \begin{itemize}
    \itemsep0em
    \item[] C.6.1. Teachers who wish to use FERed for student identification (for roll-call, exams etc.)
    \item[] C.6.2. Teachers who wish to use FERed for student surveillance
    \item[] C.6.3. Teachers who wish to use FERed for data collection
  \end{itemize}
  \item[] C.7. Detection duration preferences
  \begin{itemize}
    \itemsep0em
    \item[] C.7.1. Teachers who prefer FERed analysis at specific intervals
    \item[] C.7.2. Teachers who prefer FERed continuous analysis
  \end{itemize}
\end{itemize}

\vspace{1\baselineskip}
A first conclusion that may be drawn from this classification is that, unless persuasion and/or training efforts are invested, teachers in categories A.1.1., A.2.1., A.3.1.2.2., A.3.2., A.4.1.2., B.1.2.2., B.1.2.3., B.5.2. and C.1.2. might not to be considered potential FERed users because they would not, cannot and/or prefer not to use FER in their classroom. It is hoped that this long list would serve FERed proponents to improve the quality of their proposals, FERed critics to better assess the implication, and FERed users to formulate more informed opinions and requirements.

\vspace{2\baselineskip}

Notes \par
\vspace{1\baselineskip}
\noindent 1. For the purpose of this study a short survey was conducted on a sample of 80 teachers from Japan, Romania and Zambia, ranging from elementary to university level, asking the following question: \par
“In your teaching practice, are you also interested in student affect (how a student feels about the lesson) or only in student cognition (how much knowledge a student acquired from the lesson)?” \par
Answer choice:
\begin{itemize}
    \item[\textbullet] Yes, I am interested in what the student feels about what I taught.
    \item[\textbullet] No, I am not interested in what the student feels about what I taught. I am only interested in what he learnt from what I taught.
\end{itemize}

\noindent Only 3 teachers out of 80 answered “No”. \par
The result of this micro-investigation supports the previously posited hypothesis that there are teachers (“cognitive teachers”) who do not need or want to use a FERed. This small finding alone implies that a top-down decision to adopt such technology may hinder teacher autonomy, face resistance, squander resources etc.\par
\vspace{1\baselineskip}
\noindent 2. In the same survey, the following question was also asked: \par
“As a parent, would you agree to FER monitoring your child’s face in class?” \par
Answer choice:
\begin{itemize}
    \item[\textbullet] Yes.
    \item[\textbullet] No.
\end{itemize}

\noindent 34 out of 80 teachers answered “No”. \par
Teachers who answered “No” to this question may justifiably be considered, on the one hand, likely to reject FERed on principle, and on the other, unlikely to assume the responsibility of using FERed in their teaching practice.

\vspace{1\baselineskip}
\noindent 3. The same short survey also asked the following question: \par
“As a teacher would you choose to use FER or would you rather rely on your natural ways of detecting emotions on your students’ faces?” \par
Answer choice:
\begin{itemize}
    \item[\textbullet] Yes, I would use it.
    \item[\textbullet] No, I would rather rely on my natural ways to detect emotion on each student face.
\end{itemize}
\noindent28 out of 80 teachers answered “No” \par

\section{Conclusion, limitations and future perspectives}

If it is true that every human being has an intellectual appetite, then its discovery is a crucial moment in one’s life. If somebody or something could capture that moment, then the appetite may not be lost. Efforts to achieve this on one’s own, or with machines that simplify or fulfil this task, should be endorsed. This paper showed that knowledge specific to the field of computer science can be made clear to outsiders in terms of ends (application) and means (technology). It also identified a number of fallacies that have misguided FERed research focus away from teachers. In an attempt to provide remedy, it proposed a categorisation of teacher-users based on teacher orientation, condition and preference, which further classified teacher-users into 96 categories and subcategories, each with its contrasting characteristics. Teachers and other potential users, can refer to these classification schemes to better understand FER technology and its application in education, as well as determine their user requirements. The proposed “teacher-users” category can also enable developers and other proponents to gain a broader view of teachers as FER users. This work may also be of value for reviewers and critics of FERed. 

One limitation is that the categories presented herein are consistent mainly with one taxonomy of affective educational objectives. Another problem is that, sometimes, speculation and argument do take the place of sound theory and evidence in studies on affective educational objectives, including the ones which have guided this study. Because FER is far from common in schools, and empirical data for analysis is hardly obtainable, this paper could only provide a starting point for understanding the relationship between FERed and (potential) teachers-users.

The proposed categorisation needs to be tested based on comprehensive coverage of teacher characteristics, case studies, and data on teacher experiences with FER, as they become available. The classification schemes need revision and extension as analytical models of affective educational objectives become more complex and the FER technology advances.

\section*{Acknowledgments}
This work is funded by the JSPS KAKENHI Grant-in-Aid for Early-Career Scientists, under grant \# JP20K13900. All opinions expressed in this material are solely the author’s and do not necessarily reflect the views of the author’s organization, JSPS or MEXT. Grateful thanks are due to Dr. David Alan Grier for his continuous guidance, and to Dr. Diana Laura Borza for the collaboration which enriched this study.

\bibliographystyle{unsrt} % Use the unsrt style for numbered bibliography in alphabetical order
\bibliography{fer} % Replace "references" with your .bib file name (without the extension)

\end{document}